\documentclass[lettersize,journal]{IEEEtran}
\usepackage{amsmath,amsfonts}
\usepackage{algorithmic}
\usepackage{algorithm}
\usepackage{array}
\usepackage[caption=false,font=normalsize,labelfont=sf,textfont=sf]{subfig}
\usepackage{textcomp}
\usepackage{stfloats}
\usepackage{url}
\usepackage{verbatim}
\usepackage{graphicx}
\usepackage{cite}
\usepackage{multirow}
\newtheorem{theorem}{Theorem}
\newtheorem{corollary}{Corollary}[theorem]
\usepackage{xcolor}

\def\BibTeX{{\rm B\kern-.05em{\sc i\kern-.025em b}\kern-.08em
    T\kern-.1667em\lower.7ex\hbox{E}\kern-.125emX}}
\usepackage{balance}

\begin{document}

\title{PrNet: A Neural Network for Correcting Pseudoranges to Improve Positioning with Android Raw GNSS Measurements}

\author{Xu Weng, KV Ling, and Haochen Liu
\thanks{Manuscript received April 19, 2021; revised August 16, 2021.}
\thanks{The authors are with Nanyang Technological University, Singapore 639798 Singapore (e-mail: xu009@e.ntu.edu.sg; ekvling@ntu.edu.sg; haochen002@e.ntu.edu.sg).}
}
\markboth{Journal of \LaTeX\ Class Files,~Vol.~14, No.~8, August~2021}%
{Shell \MakeLowercase{\textit{et al.}}: A Sample Article Using IEEEtran.cls for IEEE Journals}

\IEEEpubid{0000--0000/00\$00.00~\copyright~2021 IEEE}

\maketitle

\begin{abstract}
We present a neural network for mitigating biased errors in pseudoranges to improve localization performance with data collected from mobile phones. A satellite-wise Multilayer Perceptron (MLP) is designed to regress the pseudorange bias correction from six satellite, receiver, context-related features derived from Android raw Global Navigation Satellite System (GNSS) measurements. To train the MLP, we carefully calculate the target values of pseudorange bias using location ground truth and smoothing techniques and optimize a loss function involving the estimation residuals of smartphone clock bias. The corrected pseudoranges are then used by a model-based localization engine to compute locations. The Google Smartphone Decimeter Challenge (GSDC) dataset, which contains Android smartphone data collected from both rural and urban areas, is utilized for evaluation. Both fingerprinting and cross-trace localization results demonstrate that our proposed method outperforms model-based and state-of-the-art data-driven approaches. 
\end{abstract}


\section{Introduction}
\IEEEPARstart{S}{ince} the release of Android raw Global Navigation Satellite System (GNSS) measurements, precise localization using ubiquitous and portable Android smartphones has been expected to enable various exciting localization-based applications, such as precise vehicle navigation, smart management of city assets, outdoor augmented reality, and mobile health monitoring \cite{fu2020android}. However, it is difficult to keep such promise because the inferior GNSS chips and antennas mounted in mass-market smartphones lead to large pseudorange noise and bias \cite{zangenehnejad2021gnss,scargill2023ambient,li2019characteristics}. While filtering or smoothing can reduce pseudorange noise, it remains challenging to mitigate pseudorange bias that might be caused by multipath, non-line-of-sight (NLOS) propagation, modeling residuals of atmospheric delays, smartphone hardware delays, etc \cite{humphreys2016feasibility}.

\begin{figure}[!t]
    \centering
    \includegraphics[width=0.5\textwidth]{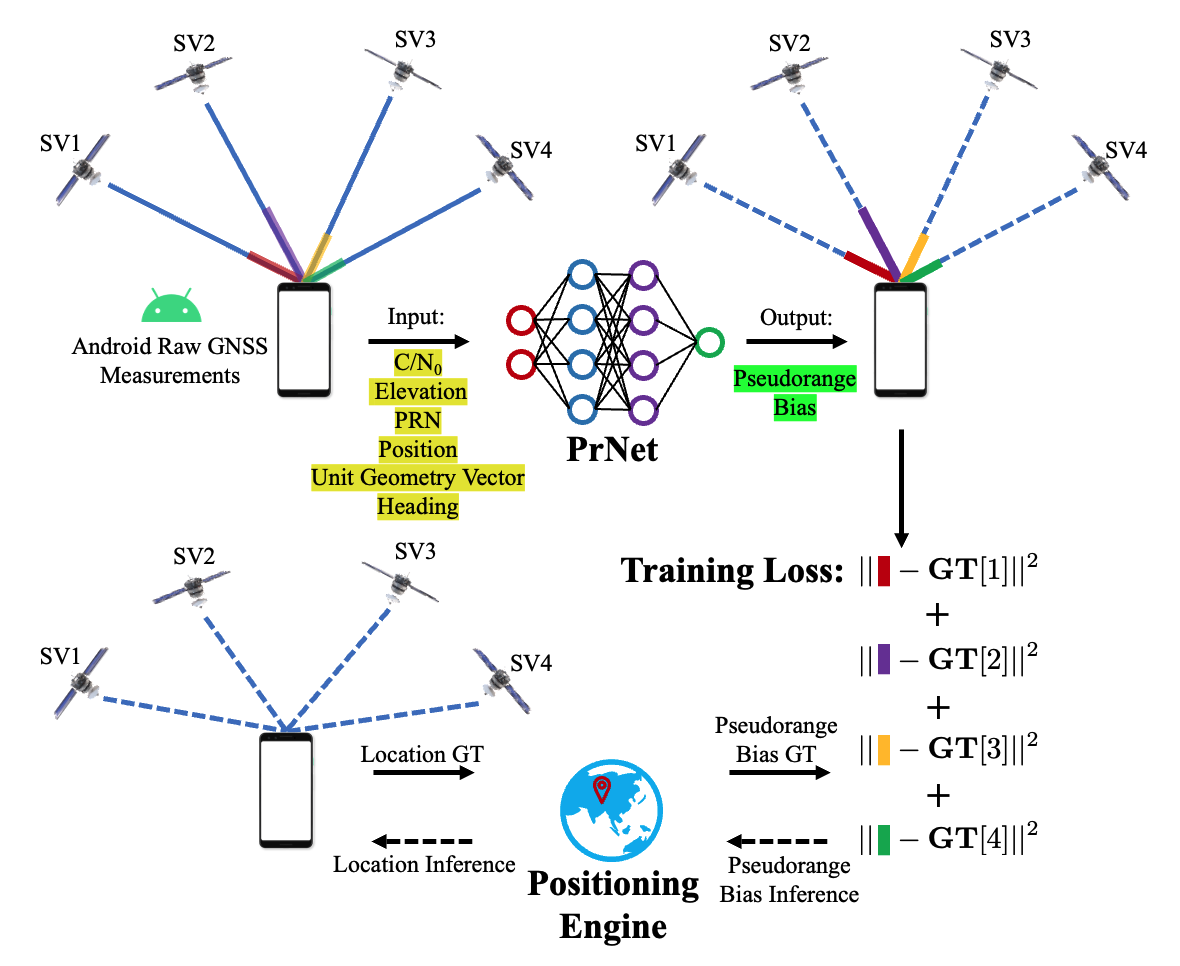}
    \caption{An overview of our PrNet-based localization pipeline. The blue solid and dashed lines represent the biased pseudoranges and unbiased pseudoranges (we assume all bias is positive here for easy illustration), respectively. The red, purple, yellow, and green segments denote the pseudorange bias of satellites SV1-SV4.}
    \label{fig: overview}
\end{figure}

To address this issue, we propose PrNet, a neural \underline{net}work for correcting \underline{p}seudo\underline{r}anges to improve positioning with Android raw GNSS measurements. As illustrated in Fig. \ref{fig: overview}, the underlying idea is to train a neural network by regressing from six satellite-receiver-context-related features to pseudorange bias. The squared loss is optimized for all visible satellites each time. After training the neural network, we can predict the biased errors, eliminate them from pseudoranges, and then feed the corrected pseudoranges into a classical localization engine to compute locations. However, we have to overcome two stumbling obstacles to implementing the above idea.\IEEEpubidadjcol

1) \emph{Feature Selection}: over 30 raw GNSS measurements are logged from Android smartphones \cite{fu2020android}, and relevant features must be carefully chosen to enhance the performance of the neural network while minimizing computational costs \cite{guyon2003introduction}. 

2) \emph{Data Labeling}: the training data should be labeled with pseudorange bias while we can only obtain the ground truth of smartphones' locations \cite{fu2020android}. Previous research has proposed various methods to estimate the pseudorange bias for geodetic GNSS receivers but ignored the pseudorange noise \cite{xu2019intelligent,sun2020improving,zhang2021prediction,sun2023resilient}. These methods degrade their own performance when transferred to Android measurements that are about one order of magnitude noisier than geodetic-quality ones.

To this end, we nominate two new features by visualizing Android measurements across various dimensions and derive the target values of pseudorange bias using location ground truth and the Rauch-Tung-Striebel (RTS) smoothing algorithm\cite{rauch1965maximum}. Besides, our experiments show that incorporating  estimation residuals of smartphone clock bias into the loss function can enhance the inference ability of the neural network. To recapitulate briefly, our contributions are:
\begin{itemize}
\item{A pipeline for learning the mapping from six satellite, receiver, context-related inputs to pseudorange bias representation, which is parameterized by a pragmatic satellite-wise Multilayer Perceptron (MLP). This includes two new proposed inputs—unit geometry vectors and smartphone headings—and a visible satellite mask layer to enable parallel computation.}
\item{Computation methods to derive labels for pseudorange bias and a differentiable loss function involving estimation residuals of smartphone clock bias.}
\item{A comprehensive evaluation from perspectives of horizontal positioning errors, generalization ability, computational load, and ablation analysis. Our codes are available at \url{https://github.com/ailocar/prnet}.}
\end{itemize}

We demonstrate that our proposed PrNet-based localization framework outperforms both the model-based and data-driven state-of-the-art pseudorange-based localization approaches. To the best of our knowledge, we present the first data-driven framework for pseudorange correction and localization over Android smartphone measurements. 

\section{Motivation}
\begin{figure}[!t]
    \centering
    \includegraphics[width=0.48\textwidth]{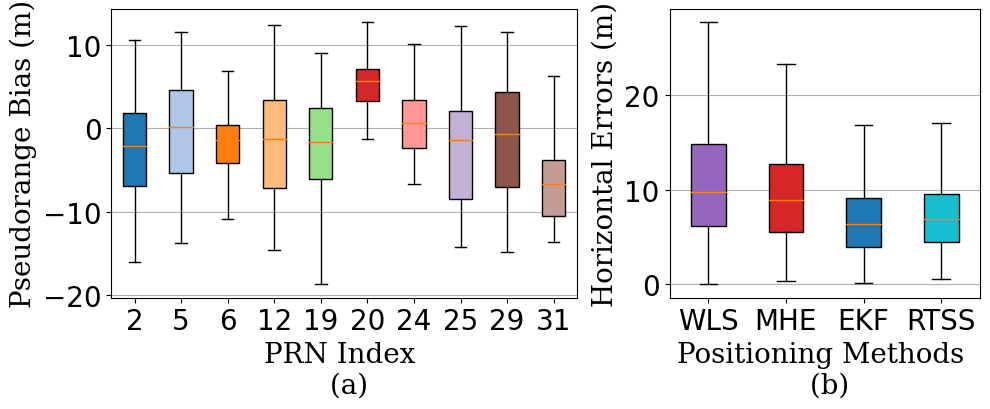}
    \caption{Impact of pseudorange bias (Trace ``2021-04-28-US-MTV-1" by Pixel4 in GSDC dataset as an example). (a) Pseudorange bias of all visible satellites throughout the trace. (b) Horizontal errors calculated with Vincenty’s formulae using Weighted Least Squares (WLS), Moving Horizon Estimation (MHE), Extended Kalman Filter (EKF), and Rauch-Tung-Striebel Smoother (RTSS). }
    \label{fig: motivation}
\end{figure}

The primary motivation behind this paper is to remove the biased errors present in pseudoranges collected from Android smartphones. To investigate the pseudorange bias and its impact on horizontal positioning accuracy, we derive the pseudorange bias as detailed in Section \ref{sec: prmBias} and adopt Weighted Least Squares (WLS), Moving Horizon Estimation (MHE), Extended Kalman Filter (EKF), and Rauch-Tung-Striebel (RTS) smoother to compute locations with the biased pseudoranges. We choose a trace of data in the Google Smartphone Decimeter Challenge (GSDC) dataset \cite{fu2020android} and display our findings in Fig. \ref{fig: motivation}. As depicted in Fig. \ref{fig: motivation}(a), biased errors are pervasive across all visible satellites and can reach magnitudes up to 10 meters. Such biased errors might be attributed to multipath, NLOS, residuals in atmospheric delay modeling, smartphone hardware interference, etc. Importantly, they are hardly modeled mathematically. Furthermore, Fig. \ref{fig: motivation}(b) clearly shows how the biased pseudoranges directly translate into localization errors that are challenging to mitigate using conventional filtering or smoothing techniques. 

\section{Related Work}
This paper focuses on artificial intelligence (AI) for GNSS localization using pseudoranges, which can be categorized into five types. \textbf{(i)} \emph{AI for Pseudorange Correction}: Recently, several learning-based methods have been proposed to predict and correct pseudorange errors using raw GNSS measurements as inputs \cite{phan2013gps,sun2020improving,sun2023resilient,zhang2021prediction}. However, all these methods use pseudorange errors (comprising noise and bias) to label training data and cannot be transferred to Android raw GNSS measurements directly. \textbf{(ii)} \emph{AI for Position Correction}: Various machine learning methods have been proposed to predict offsets between model-based position estimations and ground truth locations \cite{lan20203, caparra2021machine}. Then, the model-based position estimations are compensated with the learned location offsets. \textbf{(iii)} \emph{End-to-end Neural Networks for GNSS}: This type of work directly replaces the model-based positioning engines with deep neural networks \cite{kanhere2022improving,mohanty2022learning}. For example, authors in \cite{kanhere2022improving} leveraged a set transformer to replace the WLS engine to solve the pseudorange equations linearized about an initial location guess. Test results showed that the set transformer is sensitive to the initial location guess to a certain extent. Compared with the first type, these two kinds of approaches need to learn how to compute locations, but it has been robustly and mathematically well-established. \textbf{(iv)} \emph{AI Enhanced Localization Engine}: AI has been utilized to improve the physical models employed in conventional localization engines \cite{zhou2017least,gupta2022designing,ding2022learning}. For example, a parameter of a filter-based localization engine can be learned instead of being tweaked empirically \cite{ding2022learning}. \textbf{(v)} \emph{AI for Signal Classification}: The final category involves using AI to detect and classify multipath and NLOS propagations. By identifying pseudoranges containing multipath or NLOS errors, these methods can exclude them from the further localization process \cite{hsu2017gnss,sun2020gradient,sun2022stacking}.

\section{Preliminaries of GNSS}\label{sec:preGnss}
After the corrections of atmospheric delays as well as satellite clock offsets \cite{kaplan2017understanding}, the pseudorange measurement $\rho_{c_k}^{(n)}$ from the $n^{th}$ satellite to a smartphone at the $k^{th}$ time step is shown below.
\begin{equation}\label{eq:cpr}
\rho_{c_k}^{(n)}=r_k^{(n)}+\delta t_{u_k}+\varepsilon_k^{(n)}
\end{equation}
where the subscript $k$ represents the $k^{th}$ time step. $r_k^{(n)}$ denotes the geometry distance from the $n^{th}$ satellite to the smartphone. $\delta t_{u_k}$ represents the clock offsets of the smartphone relative to the GNSS reference time. We wrap up multipath delays, hardware delays, pseudorange noise, and other potential errors in one item $\varepsilon_k^{(n)}$ called pseudorange errors. Then, we can estimate the smartphone's location $\mathbf{x}_k = \left[x_k, y_k, z_k\right]^T$ in the Earth-centered, Earth-fixed (ECEF) coordinate system and its clock offset $\delta t_{u_k}$ by solving the following linear equation system\cite{kaplan2017understanding} established by $M$ visible satellites:
\begin{equation}\label{eq:PosEq}   
\mathbf{W}_k\mathbf{G}_k\left(\mathbf{X}_k-\tilde{\mathbf{X}}_k\right)=\mathbf{W}_k\Delta \boldsymbol{\rho}_{c_k}  
\end{equation}
where $\mathbf{X}_k=\left[x_k, y_k, z_k, \delta t_{u_k}\right]^T$ is the unknown user state while $\tilde{\mathbf{X}}_k=\left[\tilde{x}_k, \tilde{y}_k, \tilde{z}_k,\delta \tilde{t}_{u_k}\right]^T$ is an approximation of the user state. $\mathbf{W}_k$ is a diagonal matrix with the reciprocals of 1-$\sigma$ pseudorange uncertainty of all visible satellites as its main diagonal to weight pseudoranges. The geometry matrix $\mathbf{G}_k$ is calculated with the satellite location $\mathbf{x}_k^{(n)}=\left[x_k^{(n)}, y_k^{(n)}, z_k^{(n)}\right]^T$ and the approximate user location $\tilde{\mathbf{X}}_k$ \cite{kaplan2017understanding}:
\begin{equation}\label{eq:G}
    \mathbf{G}_k=\left[
                \begin{array}{clcr}
                 {a}_{x_k}^{(1)}&{a}_{y_k}^{(1)}&{a}_{z_k}^{(1)}&1\\
                 {a}_{x_k}^{(2)}&{a}_{y_k}^{(2)}&{a}_{z_k}^{(2)}&1\\
                 \dots\\
                 {a}_{x_k}^{(M)}&{a}_{y_k}^{(M)}&{a}_{z_k}^{(M)}&1 
                \end{array}\right]
\end{equation}
where,
\begin{gather}
    a_{x_k}^{(n)}=\frac{\tilde{x}_k-x_k^{(n)}}{\tilde{r}_k^{(n)}},\ a_{y_k}^{(n)}=\frac{\tilde{y}_k-y_k^{(n)}}{\tilde{r}_k^{(n)}},\ 
a_{z_k}^{(n)}=\frac{\tilde{z}_k-z_k^{(n)}}{\tilde{r}_k^{(n)}}\nonumber
\\
    \tilde{r}_k^{(n)} = \sqrt{\left(\tilde{x}_k-x_k^{(n)}\right)^2+\left(\tilde{y}_k-y_k^{(n)}\right)^2+\left(\tilde{z}_k-z_k^{(n)}\right)^2} \nonumber 
    \\
    n=1,2,3,\cdots, M \nonumber
\end{gather}
The pseudorange residual $\Delta\boldsymbol{\rho}_{c_k}$ for the $M$ visible satellites at the $k^{th}$ time step are shown as follows.
\begin{equation}
    \Delta\boldsymbol{\rho}_{c_k} = \left[\Delta \rho_{c_k}^{(1)},\Delta \rho_{c_k}^{(2)},...,\Delta \rho_{c_k}^{(M)}\right]^T \nonumber
\end{equation}
where
\begin{gather}
    \Delta \rho_{c_k}^{(n)}=\rho_{c_k}^{(n)}-\tilde{r}_k^{(n)}-\delta \tilde{t}_{u_k}-\varepsilon_k^{(n)} \nonumber
    \\
    n=1,2,3,\cdots, M \nonumber
\end{gather}

The WLS-based solution to \eqref{eq:PosEq} is shown below \cite{kaplan2017understanding}.
\begin{IEEEeqnarray}{rCl}\label{eq:xk}
    \mathbf{X}_k&=&\tilde{\mathbf{X}}_k+\Delta \mathbf{X}_k \nonumber
    \\
    &=&\tilde{\mathbf{X}}_k+\left(\mathbf{W}_k\mathbf{G}_k\right)^+\mathbf{W}_k\Delta\boldsymbol{\rho}_{c_k}
\end{IEEEeqnarray}
where $\Delta\mathbf{X}_k=\left(\mathbf{W}_k\mathbf{G}_k\right)^+\mathbf{W}_k\Delta\boldsymbol{\rho}_{c_k}$ is the displacement from the approximate user state $\tilde{\mathbf{X}}_k$ to the true one. The approximate user state $\tilde{\mathbf{X}}_k$ will be updated with the result of \eqref{eq:xk}, and the computation in \eqref{eq:xk} will be iterated until the accuracy requirement is satisfied. 

Note that the pseudorange error $\varepsilon_k^{(n)}$ is unknown in practice. The estimated user state $\hat{\mathbf{X}}_k=\left[\hat{x}_k, \hat{y}_k, \hat{z}_k,\delta \hat{t}_{u_k}\right]^T$ in the presence of pseudorange errors is:
\begin{equation}
\hat{\mathbf{X}}_k=\tilde{\mathbf{X}}_k+\Delta \mathbf{X}_k+\left(\mathbf{W}_k\mathbf{G}_k\right)^+\mathbf{W}_k\boldsymbol{\varepsilon}_k \nonumber
\end{equation} 
where,
\begin{equation}  \boldsymbol{\varepsilon}_k=\left[\varepsilon_k^{(1)},\varepsilon_k^{(2)},...,\varepsilon_k^{(M)}\right]^T\nonumber
\end{equation}

The resulting state estimation error $\mathbf{\epsilon}_{\mathbf{X}_k}$ is:
\begin{IEEEeqnarray}{rCl}\label{eq:exk} \mathbf{\epsilon}_{\mathbf{X}_k}&=&\mathbf{X}_k-\hat{\mathbf{X}}_k=-\left(\mathbf{W}_k\mathbf{G}_k\right)^+\mathbf{W}_k\boldsymbol{\varepsilon}_k \nonumber
\\
&=&\left[\epsilon_{x_k}, \epsilon_{y_k}, \epsilon_{z_k}, \epsilon_{\delta t_{u_k}}\right]^T
\end{IEEEeqnarray}

\section{Data Labeling} \label{sec: prmBias}
In this section, we set forth how to compute the target values of pseudorange bias using location ground truth and RTS smoother.

\subsection{Estimating Pseudorange Errors}
According to \eqref{eq:cpr}, pseudorange errors can be calculated as:
\begin{equation}\label{eq:PrErr}
    \varepsilon_k^{(n)} = \rho_{c_k}^{(n)}-r_k^{(n)}-\delta t_{u_k} \nonumber
\end{equation}
However, we do not have the ground truth of the smartphone's clock bias $\delta t_{u_k}$. Thus, we employ its WLS-based estimation $\delta \hat{t}_{u_k}$ to calculate pseudorange errors:
\begin{equation}\label{eq:HatPrErr}
    \hat{\varepsilon}_k^{(n)} = \rho_{c_k}^{(n)}-r_k^{(n)}-\delta \hat{t}_{u_k}
\end{equation}
Substituting \eqref{eq:cpr} and \eqref{eq:exk} into \eqref{eq:HatPrErr} yields:
\begin{IEEEeqnarray}{rCl}\label{eq:HatPrErr2}
    \hat{\varepsilon}_k^{(n)} &=& \varepsilon_k^{(n)}+\epsilon_{\delta t_{u_k}} \nonumber
    \\
    &=&\varepsilon_k^{(n)}-\mathbf{h}_k^T\boldsymbol{\varepsilon}_k
\end{IEEEeqnarray}
where $\mathbf{h}_k^T$ is the last row vector of $\left(\mathbf{W}_k\mathbf{G}_k\right)^+\mathbf{W}_k$. With \eqref{eq:HatPrErr2} as target values, we will train a neural network to regress pseudorange errors using gradient descent, which will be explained in details in Section \ref{sec: prnet}. Note that in this way, an additional shared bias might be imposed on pseudorange predictions for visible satellites, but has no impact on positioning results \cite{kaplan2017understanding}. In the following section, we will further analyze \eqref{eq:HatPrErr2} and bring up an approach to suppress the noise present in it. 

\subsection{Smoothing Pseudorange Errors}\label{sec: datasets}
The pseudorange error $\varepsilon_k^{(n)}$ couples the biased error $\mu_k^{(n)}$ and unbiased noise $v_k^{(n)}$ together:
\begin{equation}\label{eq:PrErrDecom}
    \varepsilon_k^{(n)} = \mu_k^{(n)}+v_k^{(n)}
\end{equation}
where,
\begin{gather}
   \mu_k^{(n)} = {\rm E}\left(\varepsilon_k^{(n)}\right) \nonumber
   \\
   v_k^{(n)} = \varepsilon_k^{(n)} - \mu_k^{(n)} \nonumber
\end{gather}
Substituting \eqref{eq:PrErrDecom} into \eqref{eq:HatPrErr2} yields:
\begin{IEEEeqnarray}{rCl}\label{eq:PrErr_de_rand}
    \hat{\varepsilon}_k^{(n)} &=&\mu_k^{(n)}-\mathbf{h}_k^T\mathbf{M}_k+v_k^{(n)}-\mathbf{h}_k^T\boldsymbol{v}_k 
\end{IEEEeqnarray}
where,
\begin{gather}
    \mathbf{M}_k=\left[\mu_k^{(1)},\mu_k^{(2)},\cdots,\mu_k^{(M)}\right]^T \label{eq:bias4allSv}
   \\
    \boldsymbol{v}_k = \boldsymbol{\varepsilon}_k-\mathbf{M}_k \nonumber
\end{gather}

Next, we try to extract the biased error items $\mu_k^{(n)}-\mathbf{h}_k^T\mathbf{M}_k$ from the pseudorange error estimation $\hat{\varepsilon}_k^{(n)}$ to filter out the smartphone's pseudorange noise that has been proven much larger than that of geodetic receivers \cite{li2019characteristics}.

\begin{theorem} The biased error items $\mu_k^{(n)}-\mathbf{h}_k^T\mathbf{M}_k$ are given by $\mu_k^{(n)}-\mathbf{h}_k^T\mathbf{M}_k=\mathbf{g}_k^{(n)}\left(\bar{\mathbf{x}}_{k}-{\mathbf{x}}_{k}\right)$, where $\bar{\mathbf{x}}_{k} = \left[\bar{x}_{k}, \bar{y}_{k}, \bar{z}_{k}\right]^T$ represents the mean of the WLS-based location estimation $\hat{\mathbf{x}}_k = \left[\hat{x}_{k}, \hat{y}_{k}, \hat{z}_{k}\right]^T$, and $\mathbf{g}_k^{(n)}$ is the unit geometry vector:
\begin{gather}
\mathbf{g}_k^{(n)}=\left[{a}_{x_k}^{(n)},{a}_{y_k}^{(n)}, {a}_{z_k}^{(n)}\right]^T \label{eq:g(n)}
\end{gather}
where,
\begin{equation*}
    \tilde{\mathbf{X}}_k={\mathbf{X}}_k \nonumber
\end{equation*}
\end{theorem}

\begin{IEEEproof} 
Replace the approximate state $\Tilde{\mathbf{X}}_k$ with the real state $\mathbf{X}_k$ in \eqref{eq:PosEq}. Thus, the WLS-based state estimation $\hat{\mathbf{X}}_k$ is the optimal solution to the following optimization problem:
\begin{equation}   
\min_{\hat{\mathbf{X}}_k}{||\mathbf{W}_k\mathbf{G}_k\left(\hat{\mathbf{X}}_k-\mathbf{X}_k\right)-\mathbf{W}_k\Delta \boldsymbol{\rho}_{c_k}||}^2 \nonumber  
\end{equation}

The optimal $\hat{\mathbf{X}}_k$ satisfies  
\begin{equation} \label{eq:wlsApprox}
    \mathbf{G}_k\left(\hat{\mathbf{X}}_k-\mathbf{X}_k\right)\approx\Delta \boldsymbol{\rho}_{c_k}=\boldsymbol{\rho}_{c_k}-\mathbf{r}_k-\delta t_{u_k}=\boldsymbol{\varepsilon}_k
\end{equation}
where,
\begin{gather}    \mathbf{r}_k=\left[r_k^{(1)},r_k^{(2)},\cdots,r_k^{(M)}\right]^T \nonumber
\\
\boldsymbol{\rho}_{c_k}=\left[\rho_{c_k}^{(1)},\rho_{c_k}^{(2)},\cdots,\rho_{c_k}^{(M)}\right]^T \nonumber
\end{gather}

For the $n^{th}$ satellite, the following equation can be obtained according to \eqref{eq:G}, \eqref{eq:exk} and \eqref{eq:wlsApprox}:
\begin{equation}\label{eq:wlsApproxN}  
\mathbf{g}_k^{(n)}\left(\hat{\mathbf{x}}_{k}-{\mathbf{x}}_{k}\right)+\mathbf{h}_k^T\boldsymbol{\varepsilon}_k=\varepsilon_k^{(n)}
\end{equation}

After calculating the expected values of both sides of \eqref{eq:wlsApproxN} and rearranging the result, we have
\begin{equation}
\mu_k^{(n)}-\mathbf{h}_k^T\mathbf{M}_k=\mathbf{g}_k^{(n)}\left(\bar{\mathbf{x}}_{k}-{\mathbf{x}}_{k}\right)\nonumber
\end{equation}
\end{IEEEproof}

\begin{corollary}
The biased error items $\mu_k^{(n)}-\mathbf{h}_k^T\mathbf{M}_k$ are given by $\mu_k^{(n)}-\mathbf{h}_k^T\mathbf{M}_k=\bar{r}_k^{(n)}-{r}_k^{(n)}$, where $\bar{r}_k^{(n)}=||\bar{\mathbf{x}}_{k}-{\mathbf{x}}_k^{(n)}||_2$.
\end{corollary}

\begin{IEEEproof}
Using Taylor expansion at ${\mathbf{x}}_{k}$, we have
\begin{IEEEeqnarray}{rCl}
    \bar{r}_k^{(n)} &=& ||\bar{\mathbf{x}}_{k}-{\mathbf{x}}_k^{(n)}||_2\nonumber\\
    &\approx&||{\mathbf{x}}_{k}-{\mathbf{x}}_k^{(n)}||_2+\mathbf{g}_k^{(n)}\left(\bar{\mathbf{x}}_{k}-{\mathbf{x}}_{k}\right)\nonumber\\
    &=&{r}_k^{(n)}+\mu_k^{(n)}-\mathbf{h}_k^T\mathbf{M}_k \nonumber
\end{IEEEeqnarray}
\end{IEEEproof}

\begin{figure*}[!t]
    \centering
    \includegraphics[width=\textwidth]{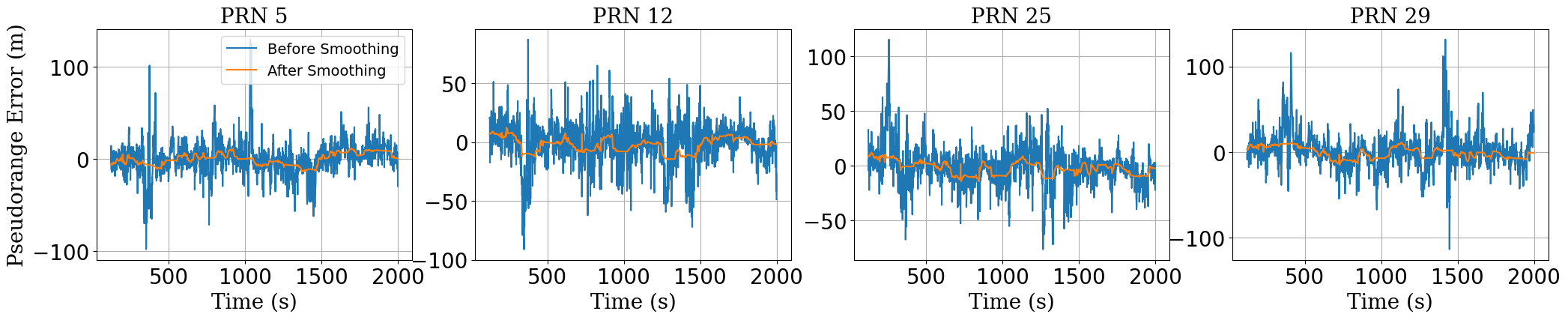}
    \caption{Pseudorange errors before and after smoothing (using four satellites from Trace ``2021-04-28-US-MTV-1" by Pixel 4 in GSDC dataset as examples)}
    \label{fig:smoothingPrErr}
\end{figure*}

This paper uses $\bar{\varepsilon}_k^{(n)} =\bar{r}_k^{(n)}-{r}_k^{(n)}$ to label training data. We consider the RTS smoother-based positioning solution as $\bar{\mathbf{x}}_{k}$. The smoothing process of RTS smoother for Android raw GNSS measurements is detailed in \cite{xu2023localization}. Fig. \ref{fig:smoothingPrErr} displays the pseudorange errors of four visible satellites before and after smoothing, indicating that the smoothed pseudorange errors are much smoother than the original ones and can represent the biased components. Note that the RTS smoother needs about 120 epochs for convergence. Therefore, during the training process, the data of the first 120 observation epochs are discarded.

\section{Feature Selection}
To enable better representation and higher efficiency of the neural network, we meticulously select input features, including the commonly utilized ones in the community and some novel features tailored specifically for Android smartphones.

\subsection{Carrier-to-noise Density Ratio $C/N_0$}
$C/N_0$ has been proven related to pseudorange errors \cite{sun2020improving,sun2023resilient,zhang2021prediction}. The $C/N_0$ of an Android smartphone is generally smaller than 50 dB$\cdot$Hz and can be normalized accordingly: 
\begin{equation}\label{eq:CN0}
    F_{1_k}^{(n)} = \frac{\rm Cn0DbHz}{50} \nonumber
\end{equation}
where ``Cn0DbHz" is one Android raw GNSS measurement.

\subsection{Elevation Angles of Satellites}
Satellite elevation angles are also closely correlated with pseudorange errors \cite{sun2020improving,sun2023resilient,zhang2021prediction}. We estimate the elevation angle $E_k^{(n)}$ of the $n^{th}$ satellite using the WLS-based positioning solution $\hat{\mathbf{x}}_k$ and the satellite location $\mathbf{x}_k^{(n)}$ \cite{kaplan2017understanding}. And it is standardized to $\left[-1, 1\right]$ as follows:
\begin{gather}\label{eq:Ele}
    F_{2_k}^{(n)} = \left[\sin{E_k^{(n)}},\cos{E_k^{(n)}}\right] \nonumber
\end{gather}

\subsection{Satellite ID}
We try to predict pseudorange errors of different visible satellites indexed by ``Svid". In this work, we only consider GPS L1 signals, and the corresponding satellite ID (PRN index) ranges from 1 to 32, so we normalize it by 32. 
\begin{equation}\label{eq:PRN}
    F_{3_k}^{(n)} = \frac{\rm Svid}{32} \nonumber
\end{equation}
where ``Svid" is one Android raw GNSS measurement.

\subsection{Position Estimations}
The surrounding environment of each location possesses unique terrain characteristics, such as the distribution of buildings, mountains, tunnels, and skyways, determining the multipath/NLOS context at the site. Thus, it is reasonable to include position estimation in the input features \cite{phan2013gps,sun2023resilient}. We can use the WLS-based position estimations that is provided as knowns in the GSDC dataset. To distinguish between closely located sites, we utilize latitude and longitude with fine-grained units, i.e., degree ($\varphi_{deg_k}$), minute ($\varphi_{min_k}$), and second ($\varphi_{sec_k}$) for latitude, and degree ($\lambda_{deg_k}$), minute ($\lambda_{min_k}$), and second ($\lambda_{sec_k}$) for longitude, which can be standardized as follows:
\begin{equation}\label{eq:WLSPosition}
    F_{4_k} = \left[\frac{\varphi_{deg_k}}{90},\frac{\varphi_{min_k}}{60},\frac{\varphi_{sec_k}}{60},\frac{\lambda_{deg_k}}{180},\frac{\lambda_{min_k}}{60},\frac{\lambda_{sec_k}}{60} \right] \nonumber
\end{equation}

\subsection{Unit Geometry Vectors}
\begin{figure*}[!t]
    \centering
    \includegraphics[width=\textwidth]{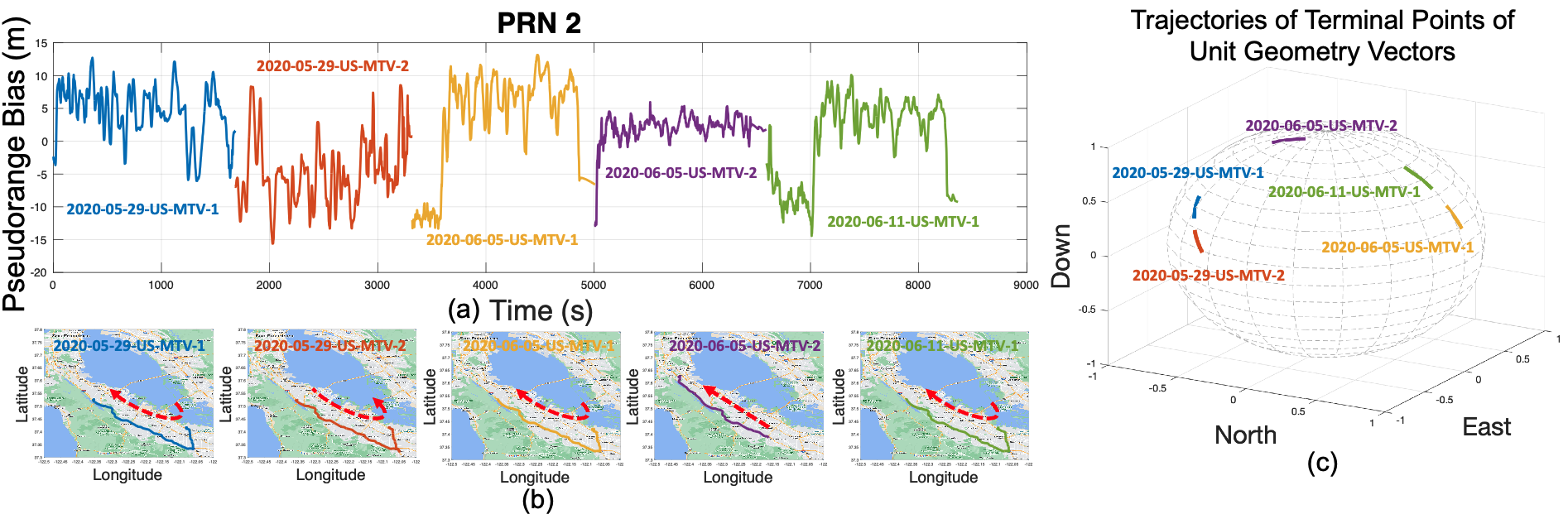}
    \caption{(a) Pseudorange bias of satellite PRN 2 along the traces collected by Pixel 4 in GSDC dataset. Different traces are plotted in a single figure to facilitate easy comparison while the time axis does not correspond to the exact moments when the data were collected; (b) Traces and directions along which each data file is collected; (c) Trajectories of terminal points of the unit geometry vectors from satellite PRN 2 to Pixel 4 on the unit sphere centered at Pixel 4.}
    \label{fig: assessment1}
\end{figure*}

The direction of signal propagation can be depicted by the unit geometry vector from the satellite to the smartphone, which is periodic at each site \cite{phan2013gps}. We visualize the unit geometry vectors in Fig. \ref{fig: assessment1}(c). Fig. \ref{fig: assessment1}(b) shows that the blue, yellow, and green traces were captured along similar routes and directions, but Fig. \ref{fig: assessment1}(a) indicates that only the yellow and green traces share similar patterns of pseudorange bias. It is on the grounds that the unit geometry vectors of yellow and green traces are close to each other but far away from the blue one, as displayed in Fig. \ref{fig: assessment1}(c). Hence, pseudorange bias are tightly correlated to unit geometry vectors. 

We convert the unit geometry vector $\mathbf{g}_k^{(n)}$ from the ECEF coordinate system to the north-east-down (NED) coordinate system to couple it to the location information (WLS-based). Each item in the unit vector falls within $\left[-1,1\right]$ and can be directly used as input features:
\begin{equation}
    F_{5_k}^{(n)} = {\mathbf{g}_{{\rm NED}_k}^{(n)}}^T\nonumber
\end{equation}

\subsection{Heading Estimations}
Fig. \ref{fig: assessment1}(b) shows that the blue and orange traces were collected along similar routes. And Fig. \ref{fig: assessment1}(c) indicates that their unit geometry vectors are also close to each other. Their pseudorange bias, however, are quite different, as shown in Fig. \ref{fig: assessment1}(a). It might be caused by the opposite moving directions along which the two data files are collected. According to the setup of Android smartphones \cite{fu2020android}, the moving direction determines the heading of smartphones' antennas, which may affect the GNSS signal reception. Therefore, we include smartphone headings $\theta_k$ into the input features, which can be approximately represented by the unit vector pointing from the current smartphone's location to the next one:
\begin{equation}
    \theta_k = \frac{\hat{\mathbf{x}}_{k+1}-\hat{\mathbf{x}}_k}{||\hat{\mathbf{x}}_{k+1}-\hat{\mathbf{x}}_k||_2} \nonumber
\end{equation}

Next, we convert the smartphone heading $\theta_k$ from the ECEF coordinate system to the NED coordinate system constructed at the current location $\hat{\mathbf{x}}_k$ and get $\theta_{{\rm NED}_k}$. Each item in the unit vector $\theta_{{\rm NED}_k}$ falls within $\left[-1,1\right]$. Thus, it can be directly included as an input feature.
\begin{equation}
    F_{6_k} = {\theta_{{\rm NED}_k}}^T \nonumber
\end{equation}

To sum up, we choose the following input features:
\begin{equation}
    \mathbf{F}_k^{(n)} = \left[F_{1_k}^{(n)}, F_{2_k}^{(n)}, F_{3_k}^{(n)}, F_{4_k}, F_{5_k}^{(n)}, F_{6_k}\right]^T \nonumber
\end{equation}
where $F_{1_k}^{(n)}$, $F_{2_k}^{(n)}$, $F_{3_k}^{(n)}$, and $F_{5_k}^{(n)}$ vary across different satellites while $F_{4_k}$ and $F_{6_k}$ are common features that are shared among all visible satellites at a given observation epoch.

\section{PrNet}\label{sec: prnet}
\begin{figure}
    \centering
    \includegraphics[width=0.48\textwidth]{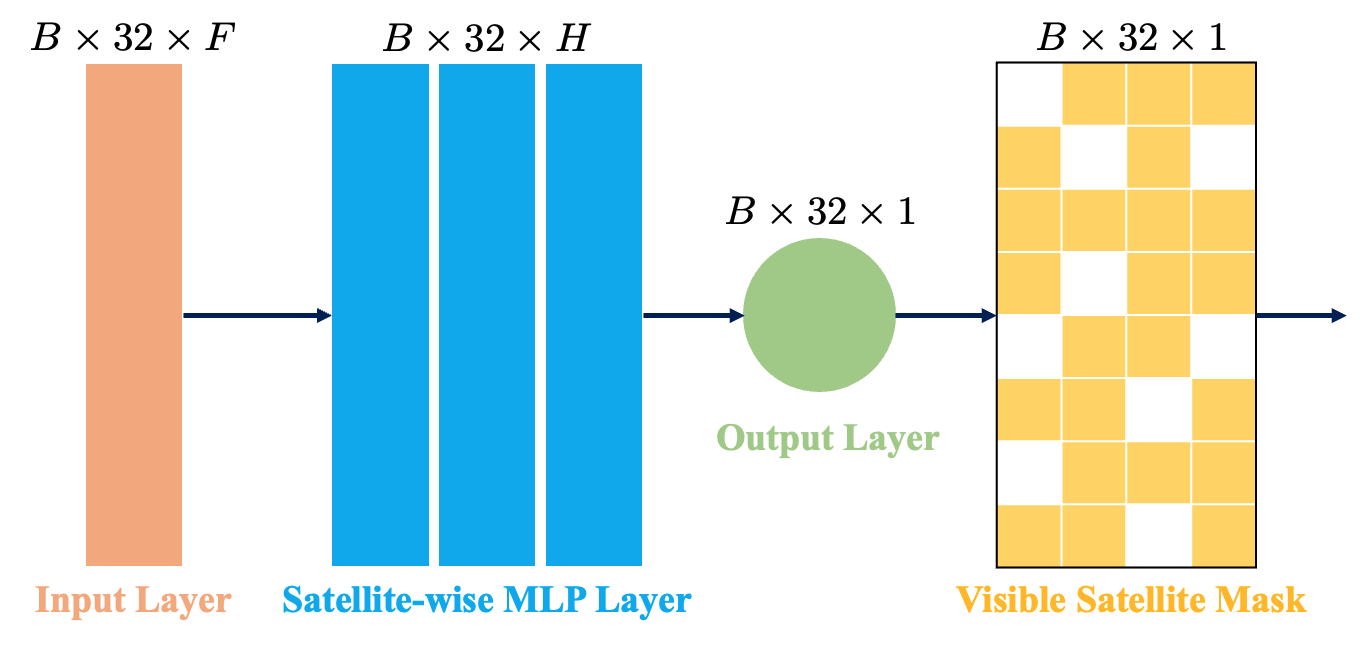}
    \caption{Diagram of PrNet. PrNet comprises a satellite-wise MLP and a visible satellite mask; $B$ represents the batch size; $F$ denotes the dimension of input features; $H$ is the number of hidden neurons.}
    \label{fig:PrNet}
\end{figure}

The MLP has proved itself in learning high-dimensional representation for regression or classification as a single network \cite{mildenhall2021nerf} or a sublayer module \cite{vaswani2017attention}. The proposed PrNet is based on a deep MLP that learns the mapping from six satellite-receiver-context-related features to pseudorange bias, i.e., ${\mu}_k^{(n)} = f\left(\mathbf{F}_k^{(n)}\right)$. The diagram of PrNet is shown in Fig. \ref{fig:PrNet}. Our approach involves passing a batch of inputs through the neural network to compute the corresponding pseudorange bias. Each sample in the batch represents all visible satellites at a given time step, and all the satellites are processed by the same MLP, called satellite-wise MLP. 

To address the challenge of varying-number satellites at different time steps, we compute the pseudorange bias of all 32 satellites in the GPS constellation each time, where the inputs of non-visible satellites are set to zero. After the output layer, we add a ``visible satellite mask" to filter out the meaningless output of non-visible satellites and retain the information of only visible satellites. This approach allows us to execute parallel computation on inputs of varying quantities.

PrNet is designed to learn the representation of the pseudorange bias $\mu_k^{(n)}$. However, the training data are labeled by $\bar{\varepsilon}_k^{(n)}=\mu_k^{(n)}-\mathbf{h}_k^T\mathbf{M}_k$. To account for this, we add the estimation residual of smartphone clock bias $-\mathbf{h}_k^T\hat{\mathbf{M}}_k$ in the loss function to align the output of PrNet and the target value $\bar{\varepsilon}_k^{(n)}=\mu_k^{(n)}-\mathbf{h}_k^T\mathbf{M}_k$:
\begin{equation}\label{eq:lossFun}
    \mathcal{L} = \sum_{n=1}^M||f\left(\mathbf{F}_k^{(n)}\right)-\mathbf{h}_k^T\hat{\mathbf{M}}_k-\bar{\varepsilon}_k^{(n)}||^2 \nonumber
\end{equation}
where,
\begin{gather}
    \hat{\mathbf{M}}_k=\left[\hat{\mu}_k^{(1)},\hat{\mu}_k^{(2)},\cdots,\hat{\mu}_k^{(M)}\right]^T \nonumber
   \\
   \hat{\mu}_k^{(n)} = f\left(\mathbf{F}_k^{(n)}\right) \nonumber
\end{gather}

\section{Implementation Details}
\subsection{Dataset}
\begin{figure*}[!t]
\centering
\subfloat[]{\includegraphics[width=0.24\textwidth]{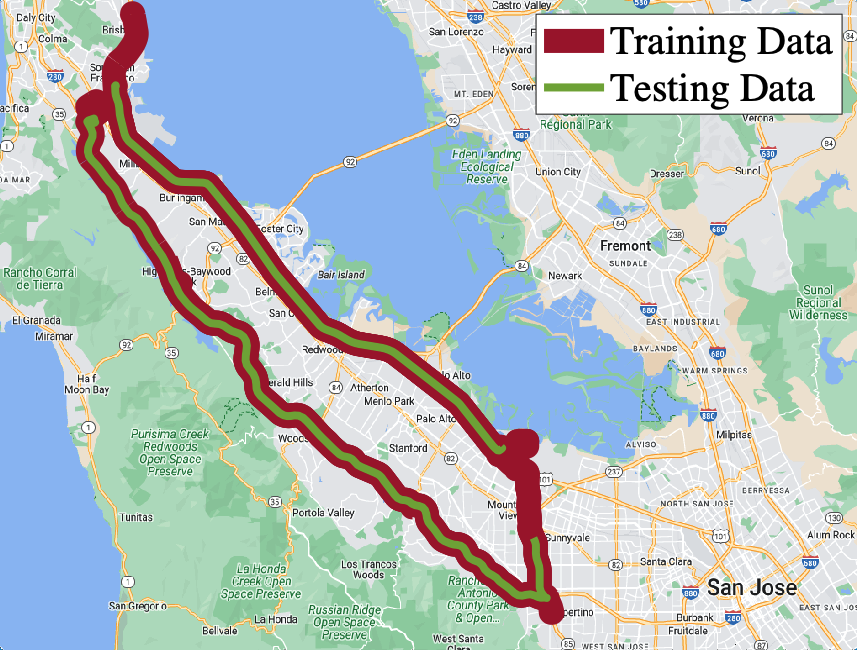}\label{fig: Scenario1}}
\hfil
\subfloat[]{\includegraphics[width=0.24\textwidth]{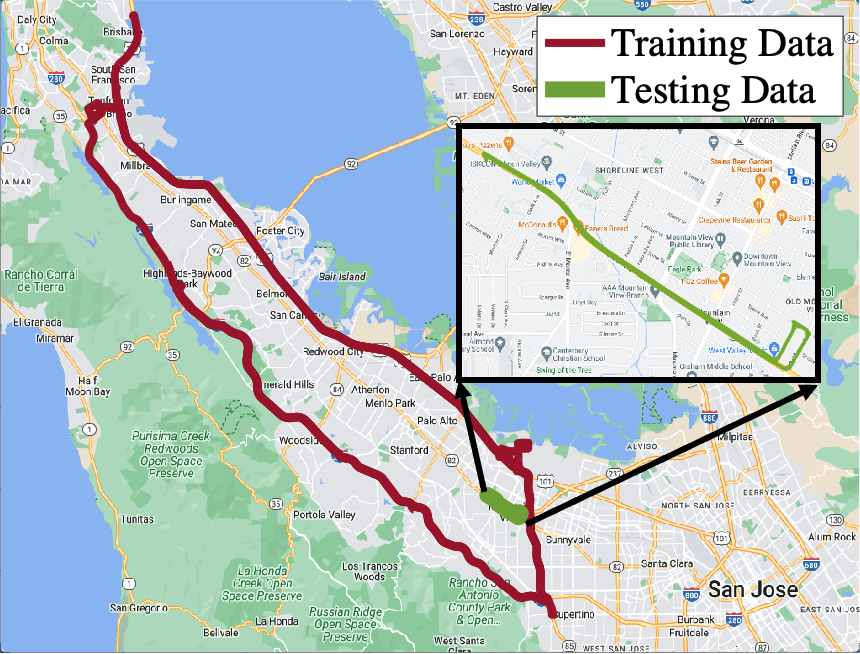}\label{fig: Scenario2}}
\hfil
\subfloat[]{\includegraphics[width=0.24\textwidth]{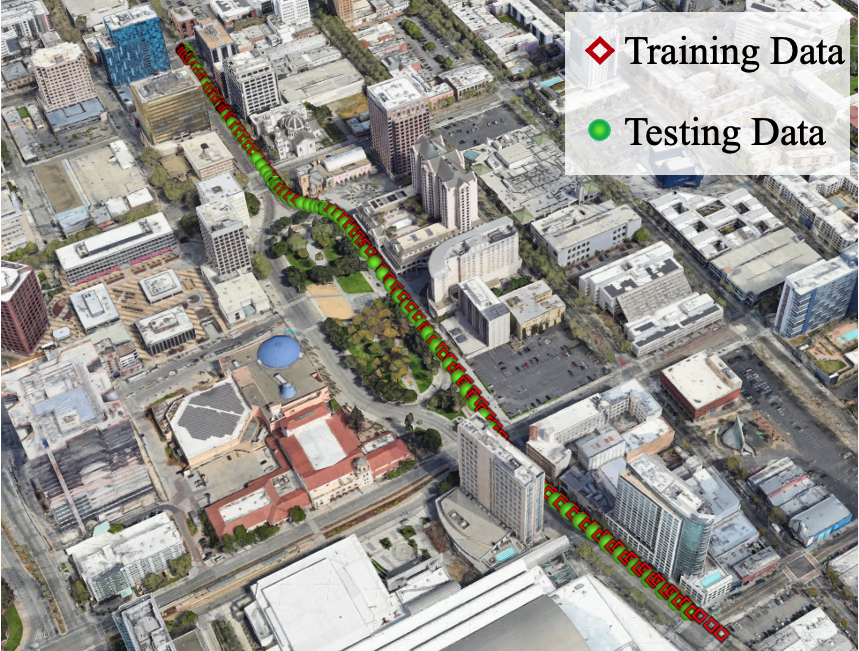}\label{fig: Scenario3}}
\hfil
\subfloat[]{\includegraphics[width=0.24\textwidth]{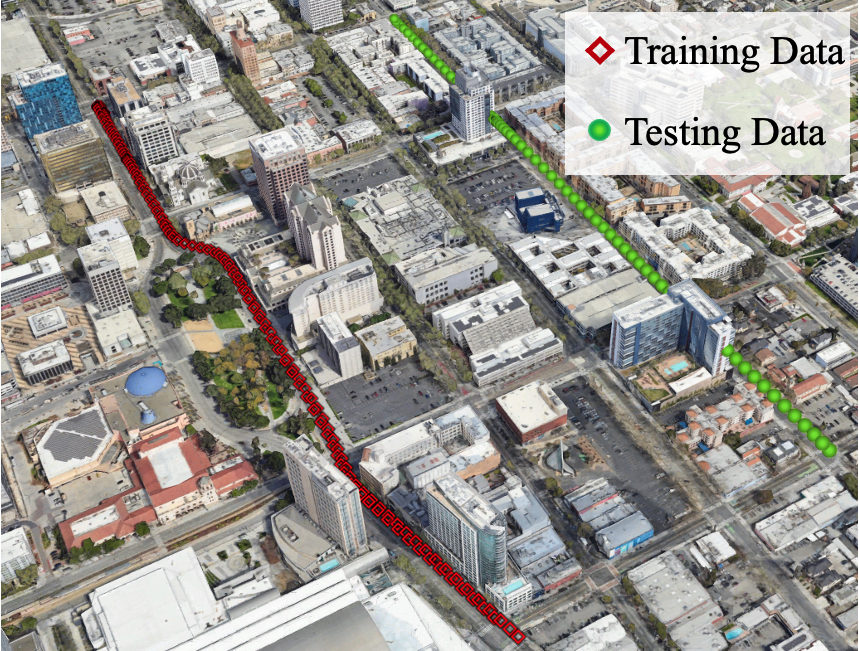}\label{fig: Scenario4}}
\caption{(a) Scenario I: rural fingerprinting. (b) Scenario II: rural cross-trace. (c) Scenario III: urban fingerprinting. (d) Scenario IV: urban cross-trace.}
\label{fig:Scenarios}
\end{figure*}

We conduct extensive evaluations on the GSDC 2021 dataset \cite{fu2020android}\footnote{We opted not to utilize the GSDC 2022 dataset due to the absence of ground truth altitude information in most of its traces, which is essential for computing the target values of pseudorange bias.}. Most of the GSDC 2021 dataset was collected in rural areas, and only a few traces were located in urban areas. As illustrated in Fig. \ref{fig:Scenarios}, we use the dataset to design four scenarios that encompass rural fingerprinting, rural cross-trace, urban fingerprinting, and urban cross-trace localization. Scenario I and Scenario II share the same training data, totaling 12 traces. Scenario III and Scenario IV share the same training data, totaling 2 traces. The training data was all collected using Pixel 4. The testing datasets for the four scenarios consist of 2, 1, 1, and 1 trace, respectively.

\subsection{Baseline Methods}
\subsubsection{Model-based methods}
We implement the vanilla WLS-based, two filtering-based (MHE and EKF), and one smoothing-based (RTS smoother) localization engines as baseline methods. More details about them can be referred to \cite{xu2023localization}.

\subsubsection{PBC-RF}
Point-based Correction Random Forest (PBC-RF) represents a state-of-the-art \emph{machine learning} approach for predicting pseudorange errors in \emph{specialized GNSS receivers}, as detailed in \cite{sun2023resilient}.

\subsubsection{FCNN-LSTM}
Fully Connected Neural Network with Long Short-term Memory (FCNN-LSTM) stands as a state-of-the-art \emph{deep learning} model designed for the prediction of pseudorange errors in \emph{specialized GNSS receivers}, as detailed in \cite{zhang2021prediction}. PBC-RF and FCNN-LSTM have not been made available as open-source software at this time. We implement them as baseline methods according to \cite{sun2023resilient,zhang2021prediction}. Our implementations are available at \url{https://github.com/Aaron-WengXu}.

\subsubsection{Set Transformer}
To the best of our knowledge, it is the only open-source state-of-the-art work that performs the data-driven art on Android raw GNSS measurements \cite{kanhere2022improving}. We trained the neural network as one baseline method. Note that its inference performance is tightly related to the initial locations of smartphones that are measured by the ``magnitude of initialization ranges" $\mu$ \cite{kanhere2022improving}. We determine the value of $\mu$ by calculating the $95^{th}$ percentile of the horizontal localization errors obtained from the model-based approaches. The set transformer we trained are available at \url{https://github.com/ailocar/deep_gnss}.

\subsection{PrNet Implementation}
The proposed neural network is implemented using PyTorch and d2l libraries \cite{zhang2021dive}. After hyper-parameter tuning, we set the number of hidden neurons $H$ to 40 and the number of hidden layers $L$ to 20. We use the Adam optimizer with a learning rate decaying from $1\times10^{-2}$ to $1\times10^{-7}$ for optimizing its weights. No regularization techniques are employed in the training process. In Scenario I and Scenario II (rural areas), the optimization of PrNet can converge within 0.5k iterations. In Scenario III and Scenario IV (urban areas), it takes about 3-5k iterations to polish up the neural network. We utilize WLS, MHE, EKF, and RTS smoother to process the pseudoranges corrected by PrNet for localization. 

\section{Experiments}

\begin{figure*}[h]
\centering
\subfloat[]{\includegraphics[width=0.246\textwidth]{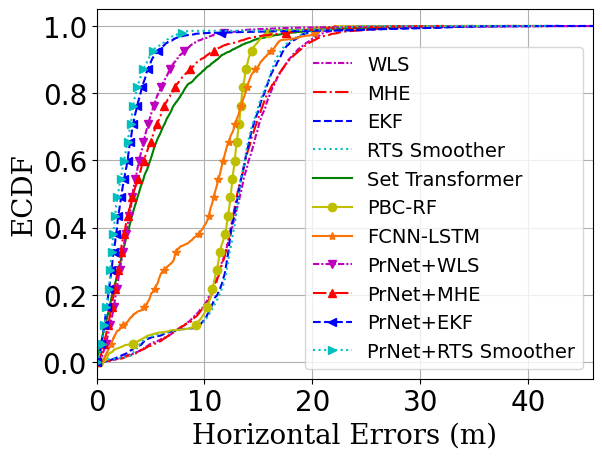}\label{fig: Results1}}
\hfil
\subfloat[]{\includegraphics[width=0.24\textwidth]{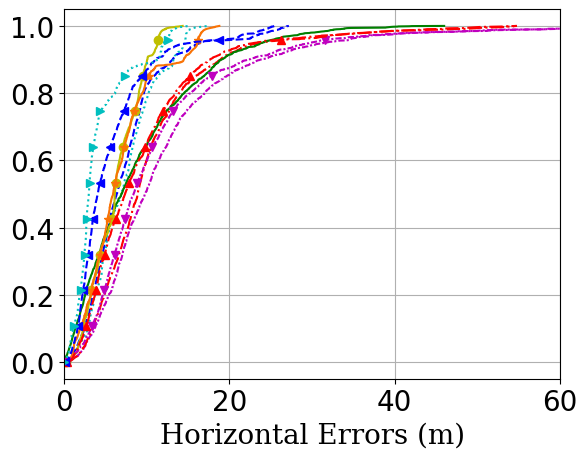}\label{fig: Results2}}
\hfil
\subfloat[]{\includegraphics[width=0.243\textwidth]{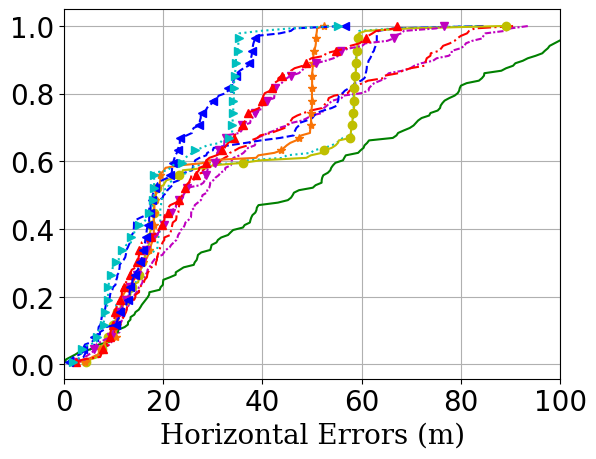}\label{fig: Results3}}
\hfil
\subfloat[]{\includegraphics[width=0.24\textwidth]{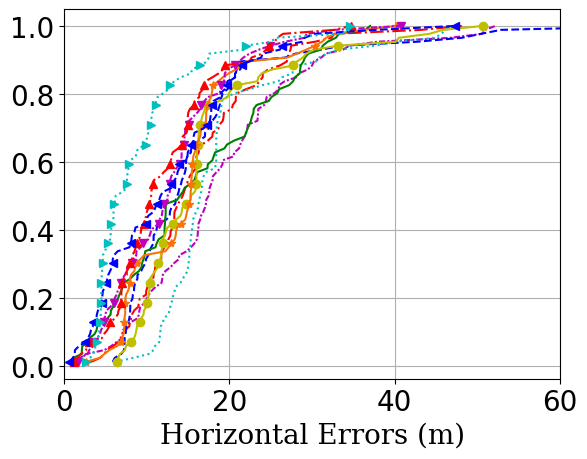}\label{fig: Results4}}
\caption{ECDF of horizontal errors in 4 scenarios. (a) Scenario I. (b) Scenario II. (c) Scenario III. (d) Scenario IV.}
\label{fig:Results}
\end{figure*}

\begin{table*}[htbp]
\caption{Horizontal positioning scores$^{\mathrm{*}}$ of different positioning methods}
\begin{center}
\begin{tabular}{c|cccc}
\hline
\multirow{2}*{\textbf{Methods}}&\multicolumn{4}{c}{\textbf{Horizontal Score (meter)}$\downarrow$}\\
&\textbf{Scenario I}&\textbf{Scenario II}&\textbf{Scenario III}&\textbf{Scenario IV}\\
\cline{1-5} 
\textbf{WLS} & 16.3901&20.6655& 54.8258& 27.2944\\
\textbf{MHE} & 16.3731&16.4572&50.9317& 22.0303\\
\textbf{EKF} & 15.5363&11.4536&41.7179& 27.1074\\
\textbf{RTS Smoother} & 15.4753&10.7628&39.5918& 27.9890\\
\textbf{Set Transformer}& 8.9388 ($\mu=15$m)&15.4394 ($\mu=15$m)&72.6576 ($\mu=60$m)& 22.9499 ($\mu=20$m)\\
\textbf{PBC-RF}$^{\mathrm{*}}$ & 13.6741&8.7351&38.8786& 26.9827\\
\textbf{FCNN-LSTM}$^{\mathrm{*}}$ & 13.8740&11.0164&34.6058& 23.1507\\
\textbf{PrNet+WLS (Ours)} & 6.2273&19.0639&43.0158& 19.4681\\
\textbf{PrNet+MHE (Ours)} & 8.1119&15.3006&41.8531& 17.9106\\
\textbf{PrNet+EKF (Ours)} & 4.4221&9.8088&28.0746& 20.0063\\
\textbf{PrNet+RTS Smoother (Ours)} & \textbf{4.0887}&\textbf{7.5405}&\textbf{26.6047}& \textbf{15.4384}\\
\hline
\multicolumn{5}{l}{$^{\mathrm{*}}$The score is the mean of the $50^{th}$ and $95^{th}$ percentile of horizontal errors calculated with Vincenty's formulae, which is}\\
\multicolumn{5}{l}{the metric adopted by Google in GSDC. Here, PBC-RF and FCNN-LSTM compute locations using RTS smoother.}
\end{tabular}
\label{tab1}
\end{center}
\end{table*}

\begin{table*}[!h]
\caption{Computational Load of Deep Learning Models}
\centering
\begin{tabular}{c|ccc|c}
\hline
\multirow{2}*{\textbf{Methods}}&\multirow{2}*{\textbf{Learnable Parameters}}&\multirow{2}*{\textbf{Computational Complexity}}&\multirow{2}*{\textbf{Sequential Operations}}&\multirow{2}*{\textbf{Inference Time$^*$ (ms)}$\downarrow$}\\
&&&&\\
\cline{1-5} 
\textbf{Set Transformer}&151107
&$\mathcal{O}(M^2)$&$2l=4$&\textbf{2.3}\\
\textbf{FCNN-LSTM}&571617&$\mathcal{O}(M)$&$M+6$&8.6\\
\textbf{PrNet}&31881
&$\mathcal{O}(M)$&$L=20$&7.5\\
\hline
\multicolumn{5}{l}{$^{\mathrm{*}}$The inference time is measured on a workstation powered by an Intel Xeon W-2133 3.6GHz CPU, 96 GB memory, and an NVIDIA TITAN V GPU.}
\\
\multicolumn{5}{l}{$M$ represents the number of visible satellites. $l$ is the number of sublayers in the encoder and decoder of the set transformer. $L$ denotes the number of}
\\
\multicolumn{5}{l}{hidden layers of PrNet.}
\end{tabular}
\label{tab2}
\end{table*}

\subsection{Horizontal Localization}
The primary focus of smartphone-based localization is on horizontal positioning errors. Therefore, we quantitatively compare our proposed frameworks against the baseline methods across four scenarios by computing their horizontal errors with Vincenty's formulae. Fig. \ref{fig:Results} displays the empirical cumulative distribution function (ECDF) of horizontal errors. The corresponding horizontal scores are summarized in Table \ref{tab1}.

The proposed PrNet+RTS smoother thoroughly outperforms all the baseline methods that employ classical model-based localization engines or sophisticated data-driven models. Additionally, by comparing the vanilla model-based approaches against their PrNet-enhanced versions, PrNet can reduce the horizontal scores by up to 74\% (PrNet+RTS smoother in Scenario I). The set transformer is tied to the ``magnitude of initialization ranges" $\mu$ and tends to yield horizontal scores around this particular value, which explains its bad performance in urban areas where $\mu$ is initialized with large magnitudes. The original intention of PBC-RF and FCNN-LSTM to process data captured from geodetic GNSS receivers limits their ability to correct pseudoranges for smartphones. Nevertheless, they outperform PrNet+WLS and PrNet+MHE in Scenario II and III. It is due to the fact that we equip them with our best localization engine (RTS smoother) that already surpasses PrNet+WLS and PrNet+MHE in these scenarios. 

\subsection{Computational Load}
Deep learning promises supreme performance at the cost of substantial computational load. We summarize the computational complexity of PrNet and the other two deep learning models, i.e., set transformer and FCNN-LSTM, in Table \ref{tab2}. In this round of competition, the set transformer outperforms the other two deep learning models even though its computational complexity is $\mathcal{O}(M^2)$. Such efficiency is credited to the transformer's parallel computation \cite{zhang2021dive} and fewer sequential operations \cite{kanhere2022improving} than the other two methods. 

\subsection{Ablation Studies}
In Table \ref{tab3}, we conduct an extensive ablation study to investigate the reliance of PrNet on our design choices, including two novel input features, loss function design, and label computation. We also assess the impact of position estimations, an input feature recently introduced in \cite{sun2023resilient}, on the resulting horizontal positioning errors. Furthermore, we analyze the scalability of PrNet by comparing models of different sizes. We present the results of PrNet+RTS smoother averaged over the four scenarios. Our best model is considered a baseline (Row 9). 

The results verify that all our design choices are reasonable. Specifically, through a stepwise removal of individual features (Rows 1-3), we observe that the two novel input features (unit geometry vectors and heading estimations) significantly impact the localization performance. In contrast, the position estimation has a trivial impact. Row 5-6 shows how the localization performance degrades if we don't involve the estimation residuals of smartphone clock bias or smoothed pseudorange errors in the loss function. Row 7 tells that PeNet can be scaled down to a smaller size with negligible performance loss, suggesting its potential for deployment on the smartphone or edge sides. Row 8 indicates that increasing the size of PrNet cannot improve its performance; in fact, it diminishes performance due to overfitting, which arises from an excessive number of learnable weights in PrNet.

\begin{table*}[!t]
\caption{Ablation studies of PrNet+RTS smoother}
\begin{center}
\begin{tabular}{l|ccc|c}
\hline
\multirow{2}*{\quad}&\multirow{2}*{\textbf{Input}}&\multirow{2}*{$\boldsymbol{(H,L)}$}&\multirow{2}*{\textbf{Loss}}&\textbf{Horizontal Score (meter)}$\downarrow$\\
&&&&\textbf{Averaged over 4 Scenarios}\\
\cline{1-5} 
\textbf{1) No Unit Geometry Vectors (UGV)}&$F_{1}F_{2}F_{3}F_{4}F_{6}$&$\left(40,20\right)$&$f\left(\mathbf{F}\right)-\mathbf{h}^T\hat{\mathbf{M}}-\bar{\varepsilon}$&17.7402\\
\textbf{2) No Heading Estimations (HE)}&$F_{1}F_{2}F_{3}F_{4}F_{5}$&$\left(40,20\right)$&$f\left(\mathbf{F}\right)-\mathbf{h}^T\hat{\mathbf{M}}-\bar{\varepsilon}$&18.3346\\
\textbf{3) No Position Estimations (PE)}&$F_{1}F_{2}F_{3}F_{5}F_{6}$&$\left(40,20\right)$&$f\left(\mathbf{F}\right)-\mathbf{h}^T\hat{\mathbf{M}}-\bar{\varepsilon}$&14.7117\\
\textbf{4) No UGV, HE, PE}&$F_{1}F_{2}F_{3}$&$\left(40,20\right)$&$f\left(\mathbf{F}\right)-\mathbf{h}^T\hat{\mathbf{M}}-\bar{\varepsilon}$&23.4167\\
\cline{1-5}
\textbf{5) No Estimation Residuals of Smartphone Clock Bias}&$F_{1}F_{2}F_{3}F_{4}F_{5}F_{6}$&$\left(40,20\right)$&$f\left(\mathbf{F}\right)-\bar{\varepsilon}$&18.1123\\
\textbf{6) No Pseudorange Error Smoothing}&$F_{1}F_{2}F_{3}F_{4}F_{5}F_{6}$&$\left(40,20\right)$&$f\left(\mathbf{F}\right)-\mathbf{h}^T\hat{\mathbf{M}}-\hat{\varepsilon}$&18.7261\\
\cline{1-5}
\textbf{7) Smaller Model}&$F_{1}F_{2}F_{3}F_{4}F_{5}F_{6}$&$\left(20,5\right)$&$f\left(\mathbf{F}\right)-\mathbf{h}^T\hat{\mathbf{M}}-\bar{\varepsilon}$& 14.6411\\
\textbf{8) Larger Model}&$F_{1}F_{2}F_{3}F_{4}F_{5}F_{6}$&$\left(40,30\right)$&$f\left(\mathbf{F}\right)-\mathbf{h}^T\hat{\mathbf{M}}-\bar{\varepsilon}$& 15.4225\\
\cline{1-5}
\textbf{9) Our Best Model}&$F_{1}F_{2}F_{3}F_{4}F_{5}F_{6}$&$\left(40,20\right)$&$f\left(\mathbf{F}\right)-\mathbf{h}^T\hat{\mathbf{M}}-\bar{\varepsilon}$&\textbf{13.4181}\\
\hline
\end{tabular}
\label{tab3}
\end{center}
\end{table*}

\subsection{Cross-phone Evaluation}
\begin{figure}[!t]
    \centering
    \includegraphics[width=0.45\textwidth]{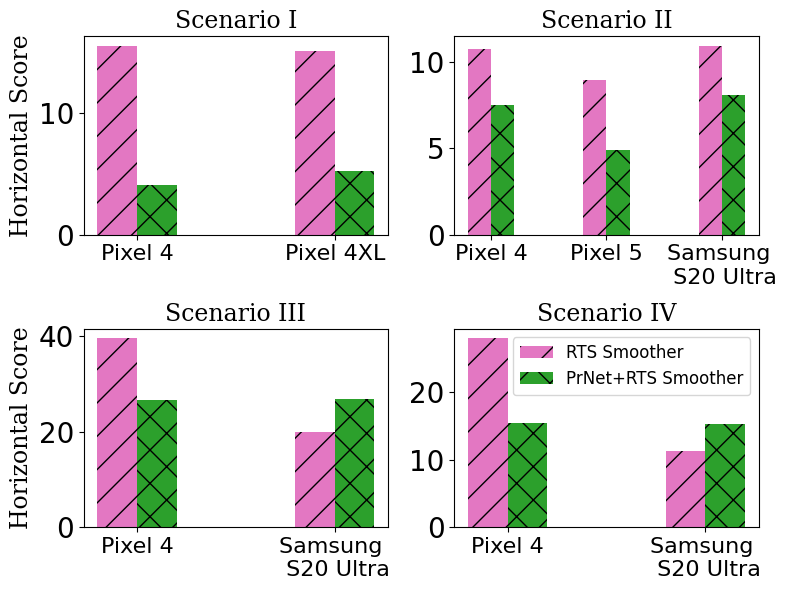}
    \caption{Cross-phone Evaluation.}
    \label{fig: CPE}
\end{figure}

To investigate the generalization ability of PrNet on various mass-market smartphones, we perform cross-phone evaluations using PrNet+RTS smoother and summarize our results in Fig. \ref{fig: CPE} (other PrNet-based methods share similar results). The training data in the four scenarios are collected by Pixel 4. During the inference process, besides the data collected by Pixel 4, we also use data from other smartphones for evaluation. Note that Google used various combinations of smartphones to collect data along different traces\footnote{In Scenario I, the data collected by Mi 8 is abnormal so that we exclude it from our analysis.}.

The results suggest that PrNet can be robustly adopted across various smartphones in rural areas (Scenario I and II), but its performance degrades when utilized on Samsung S20 Ultra in urban settings (Scenarios III and IV). Fig. \ref{fig: CPE} indicates that, in the urban areas, Samsung S20 Ultra obtains better localization performance using the standard RTS smoother compared to the performance of Pixel 4 enhanced by PrNet. Therefore, the performance gap between these two phones in urban environments leads to large data heterogeneity, which could be a potential factor behind the adaptation failures.

\section{Conclusion}
The proposed PrNet-based framework can regress the biased errors in pseudoranges collected by Android smartphones from six satellite-receiver-context-related inputs and eliminate them for better localization performance. We introduce two novel input features and meticulously calculate the target values of pseudorange bias to guide a satellite-wise MLP in learning a better representation of pseudorange bias than prior work. The comprehensive evaluation demonstrates its exceptional performance compared to the state-of-the-art approaches.

Our future work includes: 1) more advanced deep learning models can be explored to learn the satellite and temporal correlation; 2) the pseudorange-correction neural network can be trained using location ground truth in an end-to-end way; 3) more open datasets will be collected using other modalities of sensors as ground truth; 4) data augmentation strategies can be leveraged to enhance the generalization ability across heterogeneous data. 

\bibliographystyle{unsrt}
\bibliography{main}

\end{document}